\documentclass[acmsmall,screen,nonacm]{acmart}

\usepackage{booktabs}
\usepackage{subcaption}
\usepackage{graphicx}
\usepackage{pifont}

\newcommand{\cmark}{\ding{51}}
\setcopyright{none}
\renewcommand\footnotetextcopyrightpermission[1]{}
\title{TVTA: Trajectory-Aware Viseme-Guided Temporal Aggregation for Event-Based Lip Reading}

\author{Jingrong Zheng}
\affiliation{%
  \institution{Harbin Institute of Technology}
  \city{Harbin}
  \country{China}
}
\email{2023113392@stu.hit.edu.cn}

\author{Hongwei Ren}
\authornote{Corresponding author.}
\affiliation{%
  \institution{Harbin Institute of Technology}
  \city{Harbin}
  \country{China}
}
\email{ww@xxx.edu.cn}

\author{Xiangqian Wu}
\affiliation{%
  \institution{Harbin Institute of Technology}
  \city{Harbin}
  \country{China}
}

\begin{document}

\begin{abstract}
Event-based lip reading has recently emerged as a promising direction for visual speech recognition, benefiting from the high temporal resolution and motion sensitivity of event cameras. However, existing methods typically perform spatial compression before sufficient temporal modeling, which may suppress sparse and localized motion trajectories that are crucial for distinguishing similar lip movements. Moreover, most current approaches optimize temporal representations mainly at the word-classification level, leaving the underlying articulatory structure weakly constrained. To address these limitations, we propose a temporally enhanced framework for event-based lip reading. First, we introduce Trajectory-Aware Differential Aggregation (TDA), which performs local temporal modeling at each spatial location before adaptive spatial aggregation. Second, we propose Viseme-Guided Aggregation (VGA), a unified temporal module composed of a CTC decoder and a viseme-guided gated aggregation branch, which injects viseme-aware sequence supervision and improves final temporal aggregation for word recognition. Third, we incorporate an EMA teacher--student training strategy to enhance robustness under strong event perturbations. Experiments on the DVS-Lip benchmark verify the effectiveness of the proposed design, and extensive ablation studies further validate the contributions of TDA, VGA, and teacher--student consistency. Qualitative decoding results also demonstrate that the proposed CTC-based temporal modeling learns meaningful viseme-aware structure from event streams.
\end{abstract}


\maketitle

\section{Introduction}

Lip reading aims to infer spoken content from visual articulations and has become an important complement to automatic speech recognition, particularly in noisy environments and privacy-sensitive scenarios where acoustic signals are unreliable or unavailable. Compared with conventional frame-based cameras, event cameras asynchronously record brightness changes with microsecond temporal resolution, high dynamic range, and sparse event streams, making them particularly suitable for capturing subtle and rapidly evolving lip motions. These characteristics enable event cameras to preserve fine-grained visual articulations while reducing redundant appearance information, providing a promising sensing modality for visual speech recognition. Despite these advantages, effectively modeling event-based visual speech remains challenging. Unlike global facial movements, discriminative lip-reading cues are often encoded in subtle and localized articulatory patterns that evolve continuously over time. Consequently, learning temporal representations that simultaneously preserve fine-grained local dynamics and high-level articulation structures remains a fundamental challenge for event-based lip reading.

Driven by these advantages, event-based lip reading has recently attracted growing attention. The early MSTP framework~\cite{mstp} demonstrated that multigrained spatio-temporal representations can effectively model event lip streams, and its subsequent extension MSTP++~\cite{mstpp} further refined this design line. Later work explored complementary directions, including temporal granularity alignment~\cite{mtga}, event-specific triplane motion analysis~\cite{Ju2026EventbasedLR}, state-space temporal modeling~\cite{emamba}, frequency-aware enhancement~\cite{eventlip,Semantics-Aware}, and efficient neuromorphic recurrent designs~\cite{dampfhoffer2024spikgru}. These studies consistently show that event cameras provide a strong foundation for lip reading, but they also reveal that the way temporal information is organized and aggregated remains a central challenge.

Despite this progress, most existing pipelines still follow a broadly similar formulation: event streams are first converted into voxelized or structured spatial representations, then processed by a spatial encoder, and finally compressed by average pooling or related aggregation before downstream temporal reasoning and word classification~\cite{mstp,mstpp,mtga,Ju2026EventbasedLR,emamba,eventlip}. Such a design is not fully aligned with the sparse and localized nature of event data. Once spatial responses are prematurely compressed, fine-grained motion trajectories around the lips may be weakened, even though these trajectories are often precisely what distinguishes visually confusing words. Moreover, temporal optimization in existing event-based lip reading is still largely driven by the final word-level recognition objective, leaving the internal visual articulation process only weakly constrained. Although EventLip~\cite{eventlip} further introduces viseme-aware soft labels to better characterize inter-word visual similarity, its optimization is still ultimately dominated by word-level classification rather than explicit sequence-level temporal supervision.

This motivates us to revisit event-based lip reading from the perspective of temporal modeling. We argue that temporal learning should be strengthened at two complementary stages. First, local temporal evolution should be modeled before spatial aggregation, so that each spatial location can preserve its own motion trajectory rather than being immediately averaged away. Second, the final temporal representation should not rely solely on global word classification, but can also benefit from a structured intermediate sequence objective that captures visual speech progression. Recent advances in CTC-based speech recognition, including consistency-regularized CTC modeling~\cite{yao2025crctc} and intermediate-loss design for stronger sequence learning~\cite{wang2025boostingctc}, suggest that enriching CTC optimization beyond direct token prediction is a promising direction for sequence modeling.

Based on this idea, we propose a temporally enhanced framework for event-based lip reading. We first introduce Trajectory-Aware Differential Aggregation (TDA), which performs temporal modeling at each spatial location before adaptive spatial aggregation. By treating each spatial position as a temporal token sequence, TDA preserves local motion evolution prior to spatial compression and adaptively emphasizes informative responses. On top of this representation, we further propose Viseme-Guided Aggregation (VGA), a viseme-supervised temporal aggregation module that combines a CTC decoder with a viseme-guided gated aggregation branch. The CTC decoder introduces viseme-aware sequence supervision without requiring frame-level alignment, while the gated aggregation branch uses the decoded segment context to guide final temporal aggregation in the word-level temporal encoder. In addition, we incorporate an EMA teacher--student training strategy to improve robustness under strong event perturbations by enforcing consistency between weakly and strongly augmented views.

Experiments on the DVS-Lip benchmark verify the effectiveness of the proposed framework. Ablation studies further validate the contributions of TDA, VGA, and teacher--student consistency, while comparisons among Mamba, GRU, and LSTM confirm the advantage of Mamba for local temporal modeling inside the proposed TDA framework.

The main contributions of this paper are summarized as follows:
\begin{itemize}
    \item We propose Trajectory-Aware Differential Aggregation, which performs local temporal modeling before spatial aggregation and alleviates the loss of fine-grained motion cues caused by premature spatial compression in event-based lip reading.
    \item We propose Viseme-Guided Aggregation, a unified temporal module composed of a CTC decoder and a viseme-guided gated aggregation branch, which introduces viseme-aware sequence supervision and improves final temporal aggregation for word recognition.
    \item We develop an EMA teacher--student training strategy tailored to strongly perturbed event inputs and validate its effectiveness on DVS-Lip through quantitative comparisons and extensive ablation studies.
\end{itemize}

\section{Related Work}

\subsection{Event-Based Lip Reading}
\label{Event-base lip reading}

Event cameras offer microsecond temporal resolution and high dynamic range, making them ideal for capturing fast and subtle lip motions. Existing event-based lip-reading methods typically convert asynchronous streams into structured representations followed by spatio-temporal feature learning. The early MSTP framework~\cite{mstp} demonstrated the value of multi-temporal and fine-grained event representations, and its follow-up extension MSTP++~\cite{mstpp} further strengthened this design line. MTGA~\cite{mtga} then introduced graph structures to preserve local spatio-temporal details. Beyond lip reading, recent event-based recognition studies have also explored event cloud and point-based representations to better preserve sparsity and fine-grained temporal structure, including the spiking point-based architecture SpikePoint~\cite{ren2024spikepoint}, the Mamba-enhanced point framework EventMamba~\cite{ren2025eventmamba}, and the scalable event cloud network SECNet~\cite{ren2026secnet}. Recent studies in event-based lip reading have further advanced the field by integrating state-space modeling for long-range dependencies~\cite{emamba}, enhancing high-frequency lip-region details~\cite{Semantics-Aware,eventlip}, and exploring low-power neuromorphic architectures~\cite{bulzomi2023neurolip,dampfhoffer2024spikgru}. Despite these advances, most methods still follow a paradigm where temporal modeling is performed \textit{after} spatial encoding or aggregation. This design inevitably compresses localized spatial responses before they are processed temporally, which may weaken the sparse, rapidly changing motion cues that are crucial for event-based lip reading.

\subsection{Temporal Modeling and Sequence Supervision}
\label{sec:related_temporal}

Temporal context is crucial for lip reading to disambiguate similar mouth shapes. Conventional sequence backbones, such as recurrent networks~\cite{hochreiter1997lstm,cho2014gru} and Transformers~\cite{vaswani2017attention}, are effective but often suffer from sequential bottlenecks or quadratic complexity.

State Space Models (SSMs) provide an efficient alternative for long-range sequence modeling by propagating latent states with linear-time complexity~\cite{gu2022s4,gupta2022dss,smith2023s5}. Mamba~\cite{gu2023mamba} further strengthens this line by introducing input-dependent selectivity, enabling content-adaptive temporal modeling while retaining efficient hardware-aware scanning. Although SSM-based models have been extended to visual domains~\cite{zhu2024visionmamba,liu2024vmamba,li2024videomamba} and event-based lip reading~\cite{emamba}, existing methods still predominantly perform temporal modeling \textit{after} spatial encoding or aggregation. Consequently, modeling the fine-grained temporal evolution of localized spatial responses \textit{before} spatial compression remains underexplored.

\begin{figure}[t]
  \centering
  \includegraphics[width=\linewidth]{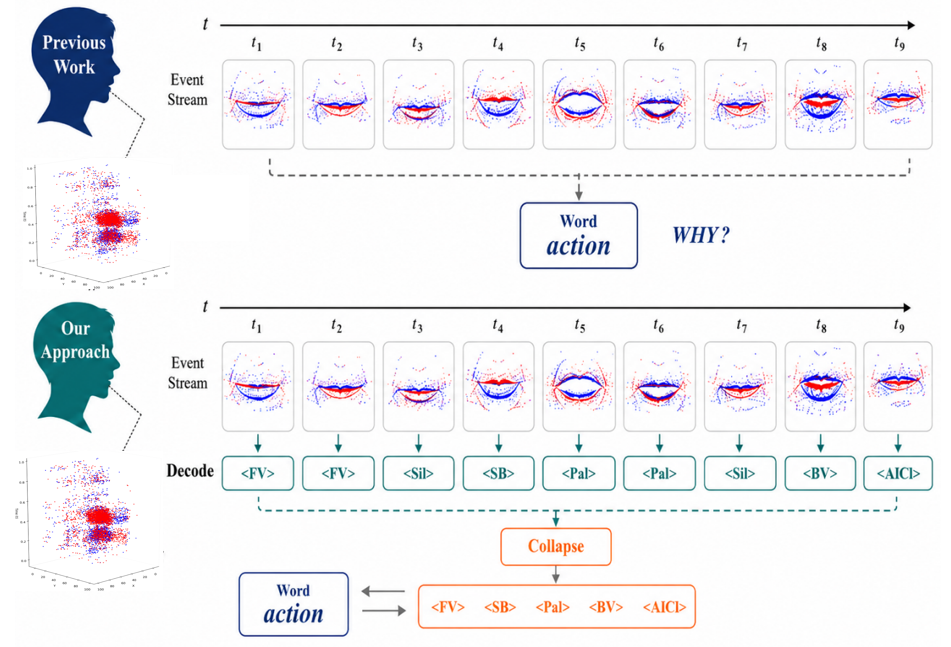}
  \caption{Comparison between previous word-level modeling and our viseme-supervised temporal decoding. Previous work directly aggregates the event stream into a word prediction, while our approach first decodes segment-level viseme labels and then collapses them into a compact viseme sequence that explains the final word recognition result.}
  \Description{A two-part comparison figure. The upper part labeled Previous Work shows an event stream directly mapped to the word action without intermediate sequence explanation. The lower part labeled Our Approach shows the same event stream decoded into segment-level viseme labels, which are then collapsed into a compact viseme sequence before producing the final word prediction.}
  \label{fig:ctc_principle}
\end{figure}

Recent visual speech recognition studies increasingly revisit temporal modeling from the perspective of sequence decoding rather than direct global classification. In continuous VSR, decoder comparisons have shown that CTC-based or hybrid sequence decoders behave differently under limited-data regimes~\cite{gimeno2024ctcattention}. VALLR~\cite{thomas2025vallr} further highlights the value of introducing an explicit intermediate sequence representation instead of relying only on direct word prediction. Beyond visual speech recognition, CR-CTC~\cite{yao2025crctc} shows that consistency regularization can also improve CTC-based sequence modeling itself. As illustrated in Fig.~\ref{fig:ctc_principle}, previous word-level methods directly aggregate event sequences into the final word prediction, whereas our approach introduces viseme-level decoding as an intermediate temporal supervision signal before the final classification stage. Existing event-based lip-reading methods still mostly formulate isolated-word recognition as global sequence aggregation followed by word classification~\cite{mstp,mstpp,mtga,emamba,eventlip}. As a result, temporal modeling is optimized mainly for word-level discrimination, while the latent articulatory structure within the sequence remains weakly constrained. This motivates us to unify two complementary temporal perspectives in a single framework: pre-aggregation temporal modeling via trajectory-aware differential aggregation, and viseme-aware sequence supervision via viseme-guided temporal aggregation.

\subsection{Teacher--Student Consistency Learning}
\label{sec:teacher_student learning}

Consistency regularization and self-distillation have emerged as powerful paradigms for improving model robustness. In self-supervised learning, representative methods such as SimCLR~\cite{chen2020simclr}, BYOL~\cite{grill2020byol}, and MoCo~\cite{he2020moco} align latent representations across different branches or differently augmented views, using mechanisms such as contrastive learning or momentum encoders to avoid representational collapse. In supervised learning, methods such as R-Drop~\cite{liang2021r} and related self-distillation approaches~\cite{furlanello2018born, zhang2019be, zhang2020self, yun2020regularizing, shen2022self} encourage prediction consistency across stochastic sub-models or perturbed branches. Related ideas have also recently been explored in CTC-based speech recognition, where CR-CTC~\cite{yao2025crctc} enforces consistency between augmented views to regularize the CTC distribution itself.

However, existing methods rarely address the vulnerability of event-based lip-reading to strong spatio-temporal perturbations. We bridge this gap by repurposing the EMA teacher-student paradigm as a fully supervised regularizer to enforce perturbation-invariant representations under event-specific augmentations.

\section{Method}

\subsection{Overall Framework}
\label{sec:overall_framework}

\begin{figure*}[t]
    \centering
    \includegraphics[width=\textwidth]{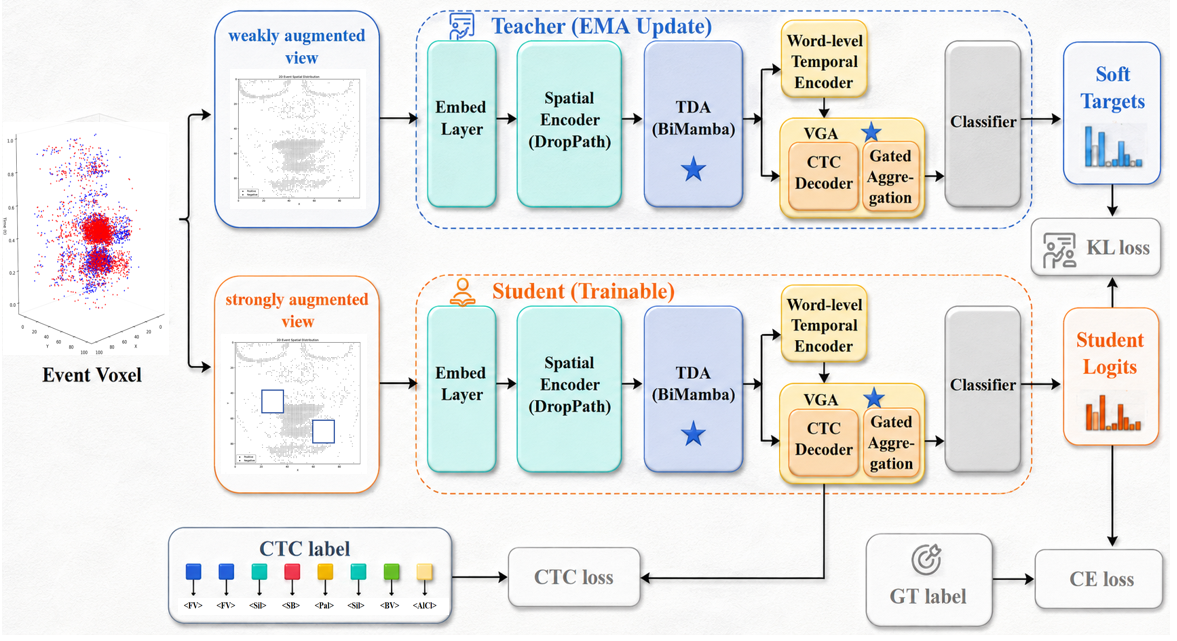}
    \caption{Overall framework of the proposed event-based lip-reading system. A weakly augmented view is processed by the EMA teacher, while a strongly augmented view is processed by the student. Both branches share the same architecture, consisting of an embedding layer, a DropPath-regularized spatial encoder, TDA, a word-level temporal encoder, VGA, and a classifier. The student is jointly optimized by cross-entropy, CTC supervision, and KL consistency, and only the student branch is used for inference.}
    \Description{The figure shows an event voxel split into weakly and strongly augmented views. The weak view is passed to an EMA teacher and the strong view to a trainable student. Each branch contains an embedding layer, a DropPath-regularized spatial encoder, TDA, a word-level temporal encoder, VGA, and a classifier. The student is supervised by cross-entropy and CTC losses, while KL consistency aligns the student with the teacher. Only the student is retained at test time.}
    \label{fig:overall_framework}
\end{figure*}

The overall architecture of the proposed event-based lip-reading framework is illustrated in Fig.~\ref{fig:overall_framework}. Given an asynchronous event stream, we first convert the events into a fixed-length dual-polarity voxel representation $\mathbf{V}\in\mathbb{R}^{T\times C_{\mathrm{in}}\times H\times W}$, where positive and negative events are accumulated in separate channels. During training, two views are generated from the same sample: a weakly augmented view for the EMA teacher and a strongly augmented view for the student.

Both branches share the same architecture. The voxel sequence is processed by an embedding layer and a DropPath-regularized ResNet encoder, followed by the proposed Trajectory-Aware Differential Aggregation module to produce frame-level features. These features are then fed into two parallel temporal branches: a word-level temporal encoder for semantic sequence modeling and the proposed Viseme-Guided Aggregation for viseme-aware temporal aggregation. VGA consists of a CTC decoder and a gated aggregation branch, where the former extracts segment-level visual speech structure and the latter uses the resulting segment context to guide the final temporal aggregation before classification.

During training, the student is optimized by word-level classification, viseme-level CTC supervision, and KL consistency against the EMA teacher. At inference time, since the teacher and student share the same model architecture and differ only in parameters, either branch can be used for performance evaluation.

\subsection{Trajectory-Aware Differential Aggregation}
\label{sec:mambapool}

The voxel input is processed by an embedding layer and a four-stage ResNet-18 spatial encoder, with DropPath~\cite{huang2016stochasticdepth} applied to the residual blocks. Instead of average pooling before temporal modeling, we introduce Trajectory-Aware Differential Aggregation to encode temporal dynamics at each spatial location and then aggregate them adaptively.

\begin{figure}[t]
  \centering
  \includegraphics[width=\linewidth]{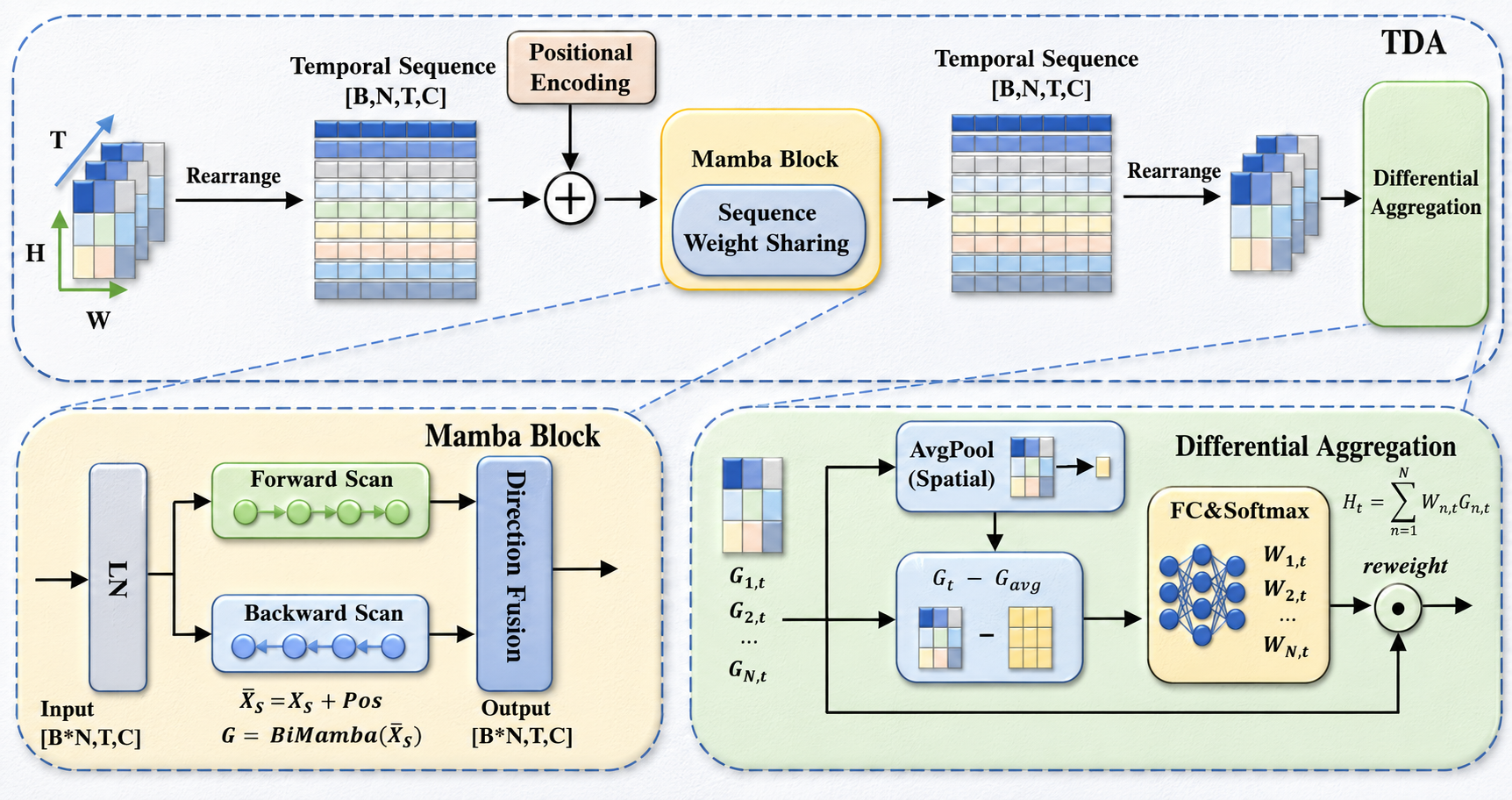}
  \caption{Overview of the proposed Trajectory-Aware Differential Aggregation module. Each spatial location is first converted into a temporal token sequence, processed by a shared bidirectional Mamba block, and then aggregated by differential spatial pooling.}
  \Description{Architecture diagram of TDA. The input feature map is rearranged into temporal sequences with positional encoding, processed by a bidirectional Mamba block with forward and backward scans plus direction fusion, and then aggregated by a differential spatial pooling module that compares raw spatial responses against the average context to produce adaptive weights.}
  \label{fig:mambapool}
\end{figure}

As shown in Fig.~\ref{fig:mambapool}, Trajectory-Aware Differential Aggregation consists of three steps: positional encoding, temporal encoding, and differential spatial aggregation. Instead of directly applying spatial pooling to the spatio-temporal feature map, we first perform temporal modeling at each spatial location so that every spatial token is equipped with temporal information before aggregation. In this way, the subsequent spatial pooling is guided by temporally informed features rather than blind spatial compression, which helps preserve motion cues that remain valuable for the downstream word-level temporal encoder. Given a high-level feature map
\begin{equation}
\mathbf{X} \in \mathbb{R}^{B \times C \times T \times H \times W},
\end{equation}
where $B$, $C$, $T$, $H$, and $W$ denote the batch size, channel dimension, temporal length, height, and width, respectively, we flatten the spatial dimensions into $N=H\times W$ spatial tokens and add a learnable spatial positional embedding to preserve spatial identity before temporal modeling:
\begin{equation}
\tilde{\mathbf{X}}_s = \mathbf{X}_s + \mathbf{P},
\quad
\mathbf{X}_s \in \mathbb{R}^{B \times N \times T \times C},
\quad
\mathbf{P} \in \mathbb{R}^{1 \times N \times 1 \times C}.
\end{equation}
Here, different spatial locations are assigned different embeddings, while the same embedding is shared across all time steps at a fixed location.
The spatial tokens are then reshaped into $(B\cdot N,T,C)$ and processed by a shared bidirectional Mamba block along the temporal dimension. As a selective state space model, Mamba~\cite{gu2023mamba} performs content-dependent temporal scanning, enabling each spatial sequence to adaptively encode temporal feature variations before spatial aggregation:
\begin{equation}
\mathbf{G} = \mathrm{BiMamba}(\tilde{\mathbf{X}}_s),
\quad
\mathbf{G} \in \mathbb{R}^{B \times N \times T \times C}.
\end{equation}
The BiMamba parameters are shared across all spatial sequences.

After temporal encoding, we compute the global spatial context for each time step:
\begin{equation}
\mathbf{G}_{\mathrm{avg}} = \frac{1}{N}\sum_{n=1}^{N}\mathbf{G}_n,
\quad
\mathbf{G}_{\mathrm{avg}} \in \mathbb{R}^{B \times 1 \times T \times C}.
\end{equation}
We then estimate the spatial aggregation weights from the differential response between each spatial token and the global spatial context:
\begin{equation}
\mathbf{W}_n =
\mathrm{Softmax}_{n}
\left(
\frac{\phi(\mathbf{G}_n-\mathbf{G}_{\mathrm{avg}})}{\tau}
\right),
\label{eq:diff_weight}
\end{equation}
where $\mathbf{G}_n \in \mathbb{R}^{B \times 1 \times T \times C}$ denotes the temporally encoded feature of the $n$-th spatial token, $\phi(\cdot)$ is an MLP that maps the differential feature to a scalar importance logit, and $\tau$ is a learnable temperature.

Finally, the frame-level temporal feature sequence is obtained by weighted spatial aggregation:
\begin{equation}
\mathbf{H}_t = \sum_{n=1}^{N} \mathbf{W}_{n,t} \mathbf{G}_{n,t},
\quad
\mathbf{H}\in\mathbb{R}^{B\times T\times C}.
\end{equation}

\subsection{Viseme-Guided Aggregation for Segment Decoding and Temporal Aggregation}
\label{sec:ctc_guided}

\begin{figure}[t]
  \centering
  \includegraphics[width=\linewidth]{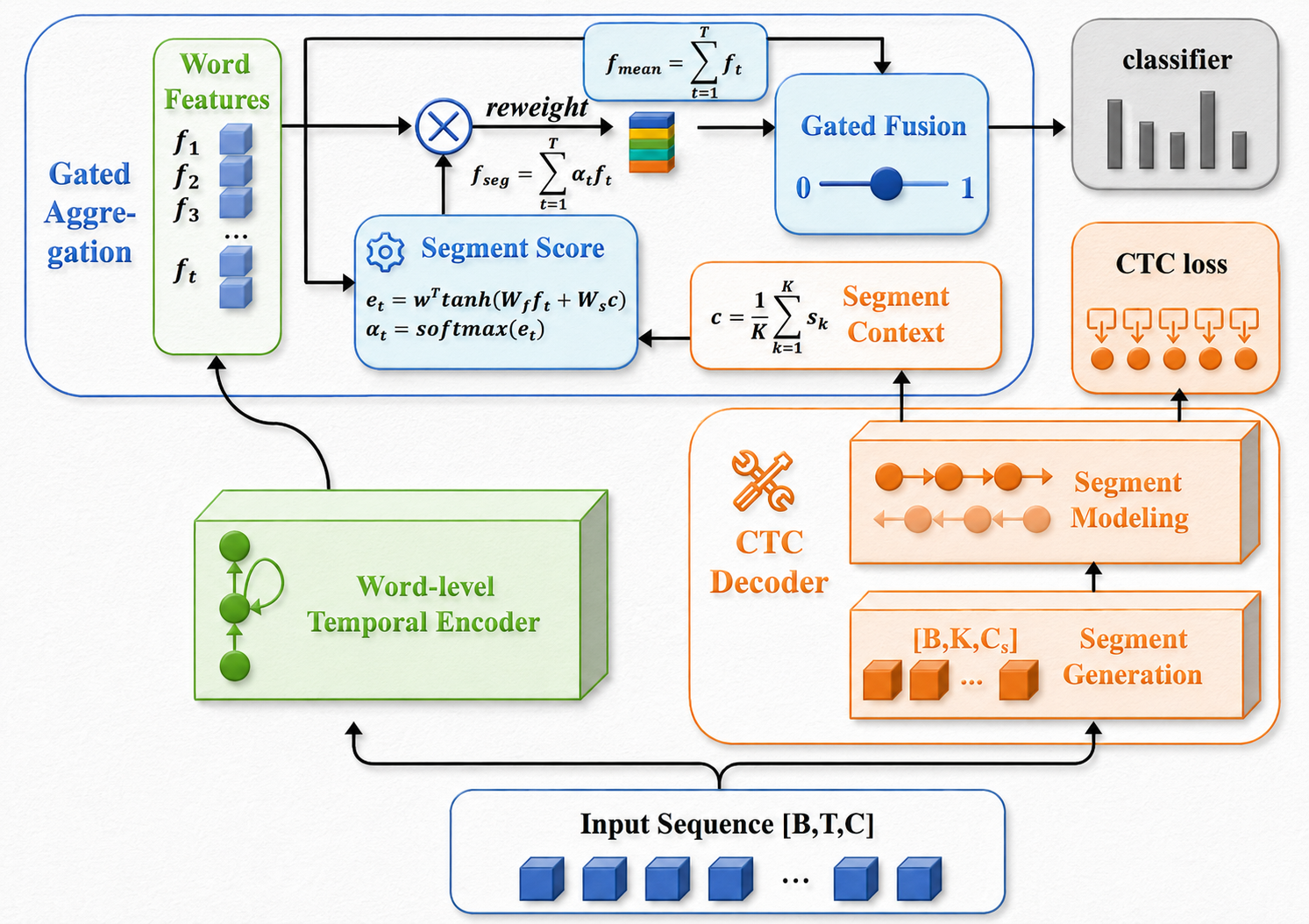}
  \caption{Overview of the proposed Viseme-Guided Aggregation. VGA consists of a CTC decoder and a gated aggregation branch. The decoder provides viseme supervision and segment context, and the gated aggregation branch uses this context to guide the final temporal aggregation of the word-level temporal encoder.}
  \Description{Architecture diagram of VGA. Frame-level features are converted into temporal segments and modeled to produce segment states. One path predicts viseme logits for CTC loss, while another averages the segment states into a segment context that guides the gated aggregation branch over the word-level temporal features before classification.}
  \label{fig:ctc_branch}
\end{figure}

The frame-level feature sequence $\mathbf{H}$ is dispatched to two parallel temporal branches: a word-level temporal encoder and the proposed Viseme-Guided Aggregation. As shown in Fig.~\ref{fig:ctc_branch}, VGA consists of two coupled parts, namely a CTC decoder for segment-level viseme supervision and a gated aggregation branch for viseme-guided temporal aggregation.

\textbf{CTC decoder.} Since DVS-Lip provides only word labels rather than frame-aligned viseme annotations, CTC is suitable because it enforces sequence structure without requiring manual alignment. Following the standard visual speech pipeline, each word label is first converted to a pronunciation sequence by a pronunciation dictionary and then mapped to a viseme sequence with a fixed conversion table following prior audio-visual speech recognition and EventLip~\cite{hazen2004segment,eventlip}. The CTC decoder first applies a segment generation module to aggregate short temporal neighborhoods into coarse segments, and then feeds them into a segment modeling module:
\begin{equation}
\mathbf{S}
=
\mathrm{SegModel}_{\mathrm{ctc}}(\mathrm{SegGen}(\mathbf{H})),
\quad
\mathbf{c}
=
\frac{1}{K}\sum_{k=1}^{K}\mathbf{S}_k,
\end{equation}
\begin{equation}
\mathbf{S}\in\mathbb{R}^{B\times K\times D_s},
\quad
\mathbf{c}\in\mathbb{R}^{B\times D_s},
\end{equation}
where $K=T/\ell$ is the number of temporal segments and $\ell$ denotes the segment length. Here, $\mathrm{SegGen}(\cdot)$ denotes the temporal convolution and segment-wise average pooling used to form coarse segment tokens. A viseme prediction head is then applied to each segment state to obtain
\begin{equation}
\mathbf{P} = \mathrm{VisemeHead}(\mathbf{S}),
\quad
\mathbf{P}\in\mathbb{R}^{B\times K\times C_{\mathrm{vi}}},
\end{equation}
where $C_{\mathrm{vi}}$ is the number of viseme categories including the CTC blank. Given the target viseme sequence associated with the word label, the decoder is trained with the standard CTC objective~\cite{graves2006ctc}:
\begin{equation}
\mathcal{L}_{\mathrm{ctc}}
=
\mathrm{CTC}
\bigl(
\log\mathrm{Softmax}(\mathbf{P}),
\mathbf{y}^{\mathrm{vi}}
\bigr),
\end{equation}
which allows the network to learn latent monotonic alignments between visual frame dynamics and viseme-level units without requiring explicit frame annotations.

\textbf{Gated aggregation.} In parallel, the word-level temporal encoder further models $\mathbf{H}$ as
\begin{equation}
\mathbf{F}
=
\mathrm{BiGRU}_{\mathrm{word}}(\mathbf{H}),
\quad
\mathbf{F}\in\mathbb{R}^{B\times T\times D_w}.
\end{equation}
where the word-level temporal encoder follows the baseline design and is implemented as a three-layer bidirectional GRU. In the baseline, the final prediction is obtained by directly classifying the temporally mean-pooled representation. Here, we retain this mean-pooled path as $\mathbf{g}_{\mathrm{mean}}$ and further introduce a viseme-guided attention-based aggregation path. Specifically, the gated aggregation branch uses the segment context from the CTC decoder to guide temporal aggregation. We then estimate temporal attention weights by jointly scoring each word-level feature and the segment context:
\begin{equation}
\alpha_t
=
\mathrm{Softmax}_t
\left(
\mathbf{w}^{\top}
\tanh
\left(
\mathbf{W}_g\mathbf{F}_t
+
\mathbf{W}_s\mathbf{c}
\right)
\right).
\end{equation}
The attention-based aggregation result and mean-pooled representation are
\begin{equation}
\mathbf{f}_{\mathrm{att}}
=
\sum_{t=1}^{T}\alpha_t\mathbf{F}_t,
\quad
\mathbf{f}_{\mathrm{mean}}
=
\frac{1}{T}\sum_{t=1}^{T}\mathbf{F}_t.
\end{equation}
We then fuse them through a learnable gate
\begin{equation}
\mathbf{r}
=
(1-\gamma)\mathbf{f}_{\mathrm{mean}}
+
\gamma\mathbf{f}_{\mathrm{att}},
\quad
\gamma=\sigma(g),
\end{equation}
where $g$ is a trainable scalar parameter. The final word logits are produced by a classifier applied to $\mathbf{r}$. In this way, VGA jointly combines viseme-aware sequence supervision and viseme-guided temporal aggregation.

\subsection{Perturbation-Consistent Teacher--Student Training}
\label{sec:ema}

We adopt an exponential moving average (EMA) teacher--student strategy following prior work~\cite{tarvainen2017mean,caron2021dino} to regularize predictions under strong event-level perturbations.

The teacher shares the same architecture as the student, but its parameters are updated by EMA:
\begin{equation}
\theta_t^{\prime}
\leftarrow
\mu(e)\theta_{t-1}^{\prime}
+
(1-\mu(e))\theta_t,
\end{equation}
where $\theta_t$ and $\theta_t^{\prime}$ denote the student and teacher parameters at training step $t$, respectively, and $\mu(e)$ is the EMA decay rate at epoch $e$.

During training, the teacher and student receive differently perturbed versions of the same sample. The teacher observes a weakly augmented input $\mathbf{x}^{w}$, while the student observes a strongly augmented input $\mathbf{x}^{s}$. Their logits are computed as
\begin{equation}
\mathbf{z}_t = f_{\theta'}(\mathbf{x}^{w}),
\quad
\mathbf{z}_s = f_{\theta}(\mathbf{x}^{s}),
\end{equation}
where $f_{\theta}$ and $f_{\theta'}$ denote the student and teacher networks, respectively. The training objective is
\begin{equation}
\mathcal{L}
 =
\lambda_{\mathrm{ce}}\mathcal{L}_{\mathrm{ce}}
+
\lambda_{\mathrm{kl}}(e)\mathcal{L}_{\mathrm{kl}}
+
\lambda_{\mathrm{ctc}}\mathcal{L}_{\mathrm{ctc}},
\end{equation}
where $\mathcal{L}_{\mathrm{ce}}$ is the cross-entropy loss, $\mathcal{L}_{\mathrm{ctc}}$ is the auxiliary CTC loss in Sec.~\ref{sec:ctc_guided}, $\lambda_{\mathrm{ce}}$ and $\lambda_{\mathrm{ctc}}$ are balancing weights, and $\lambda_{\mathrm{kl}}(e)$ is the epoch-dependent consistency weight. The KL consistency loss is
\begin{equation}
\mathcal{L}_{\mathrm{kl}}
=
D_{\mathrm{KL}}
\left(
\mathrm{Softmax}(\mathbf{z}_t)
\;\|\;
\mathrm{Softmax}(\mathbf{z}_s)
\right).
\end{equation}
The teacher prediction is detached during optimization.

Following Mean Teacher~\cite{tarvainen2017mean}, we gradually increase the KL consistency weight during training:
\begin{equation}
\lambda_{\mathrm{kl}}(e)=\lambda_{\max}g(e),
\end{equation}
where $\lambda_{\max}$ is the maximum consistency weight and $g(e)$ is a ramp-up function that increases from 0 to 1.

We use a stage-wise EMA decay schedule:
\begin{equation}
\mu(e)=
\begin{cases}
\mu_{\mathrm{small}}, & e \leq e_{\mathrm{ramp}}, \\
\mu_{\mathrm{large}}, & e > e_{\mathrm{ramp}},
\end{cases}
\end{equation}
where $e_{\mathrm{ramp}}$ denotes the ramp-up boundary and $\mu_{\mathrm{small}}<\mu_{\mathrm{large}}$. A smaller EMA decay in the early stage allows the teacher to follow the supervised student more rapidly while the student is still learning basic discriminative knowledge. After the ramp-up stage, a larger EMA decay makes the teacher evolve more smoothly, thereby providing stabler soft targets for consistency regularization.

\section{Experiments}

In this section, we describe the experimental settings, implementation details, comparison with existing methods, and ablation analysis. All experiments were conducted on a single NVIDIA GeForce RTX 3090 GPU with 24~GB memory.

\subsection{Dataset Settings}

\textbf{DVS-Lip Dataset}. DVS-Lip was introduced by MSTP~\cite{mstp} as the first event-based lip-reading dataset. It contains 19,871 word-level samples captured by a DAVIS-346 event camera from 40 volunteers. The vocabulary consists of 100 words selected from LRW~\cite{lrw}, including visually similar word pairs and common words. Each sample records the event stream generated when a volunteer articulates a target word, together with its word label.

We follow the official split of DVS-Lip, using 14,896 samples for training and 4,975 samples for testing. The official test set is divided into two subsets. The first subset, denoted as Acc1, contains visually confusing word pairs and is therefore more challenging. The second subset, denoted as Acc2, contains the remaining common words. We report Acc1, Acc2, and the overall accuracy Acc for comparison.

\subsection{Implementation Details}

\textbf{Data augmentation}. The spatio-temporal augmentations used in our experiments follow the standard training configuration adopted in prior event-based lip-reading research~\cite{dampfhoffer2024spikgru}, including random cropping, horizontal flipping, spatial cutout, random zooming, and temporal masking. During training, the mouth region is first cropped to $96\times96$, followed by a random $88\times88$ crop. Horizontal flipping is applied with a probability of $0.5$. During testing, we use a deterministic center crop to obtain the $88\times88$ input.

In our framework, these augmentations are organized asymmetrically for EMA teacher--student training. Given the same event stream, we construct a weakly augmented view for the teacher and a strongly augmented view for the student. The teacher input only uses the basic crop and horizontal flip operations, while the student input is further perturbed by spatial cutout, random zooming, and temporal masking.

\textbf{Training settings}. The model is trained on the official DVS-Lip training split for 100 epochs with a batch size of 32. We use Adam~\cite{adam} as the optimizer with a weight decay of $1\times10^{-4}$, while batch normalization parameters are excluded from weight decay. The learning rate is warmed up during the first epoch to the maximum value of $3\times10^{-4}$, and then annealed by a cosine schedule to $5\times10^{-6}$.

DropPath is applied to the residual branch of each ResBlock. The drop probabilities are increased stage by stage and set to $0.05$, $0.10$, $0.15$, and $0.20$ for the four residual stages, respectively.

For VGA, the segment length in the CTC decoder is set to $\ell=5$, and the hidden dimension is set to 1024. In the gated aggregation branch, the attention hidden dimension is set to 512, and the gate parameter is initialized to favor mean pooling at the beginning of training. The influence of the CTC balancing weight $\lambda_{\mathrm{ctc}}$ is further analyzed in the ablation study.

\textbf{Teacher--student configuration}. As described in Sec.~\ref{sec:ema}, the consistency weight is controlled by a ramp-up function $g(e)$:
\begin{equation}
g(e)=
\begin{cases}
\dfrac{
\exp\left(-5(1-x)^2\right)-\exp(-5)
}{
1-\exp(-5)
}, & e \leq e_{\mathrm{ramp}}, \\[4mm]
1, & e > e_{\mathrm{ramp}},
\end{cases}
\quad
x=
\dfrac{e-1}{e_{\mathrm{ramp}}-1}.
\end{equation}
Here, $e_{\mathrm{ramp}}$ is set to 40\% of the total training epochs. The EMA decay rate is set to $\mu=0.999$ in the early stage and switched to $\mu=0.9998$ after the ramp-up period. The influence of the KL consistency weight is further analyzed in the ablation study. The reported test performance is based on the student model unless otherwise specified.

\subsection{Main Results}

\begin{table}[t]
  \centering
  \caption{Performance comparison on the DVS-Lip dataset.}
  \label{tab:comparison}
  \begin{tabular}{@{}lccc@{}}
    \toprule
    \textbf{Model} & \textbf{Acc1 (\%)} & \textbf{Acc2 (\%)} & \textbf{Acc (\%)} \\
    \midrule
    Event Clouds~\cite{wang2019eventclouds}  & 35.82 & 48.51 & 42.15 \\
    EV-Gait-3DGraph~\cite{wang2019evgait}    & 26.35 & 37.75 & 32.04 \\
    EST~\cite{gehrig2019est}                 & 40.91 & 56.45 & 48.66 \\
    I3D~\cite{carreira2017i3d}               & 58.24 & 77.68 & 67.94 \\
    TANet~\cite{liu2021tam}                  & 58.36 & 79.17 & 68.74 \\
    ACTION-Net~\cite{wang2021actionnet}      & 58.32 & 79.41 & 68.84 \\
    AGCN~\cite{jiang2023agcn}                & 55.52 & 80.47 & 67.74 \\
    GET~\cite{peng2023get}                   & 58.96 & 80.82 & 69.80 \\
    \midrule
    MSTP~\cite{mstp}                         & 62.17 & 82.07 & 72.10 \\
    Spiking MSTP~\cite{bulzomi2023neurolip} & --    & --    & 60.20 \\
    E-Mamba~\cite{emamba}                   & 63.75 & 82.71 & 73.23 \\
    SpikingGRU2+~\cite{dampfhoffer2024spikgru} & -- & --    & 75.30 \\
    SSE-Net~\cite{ssenet}                   & 63.06 & 84.45 & 73.73 \\
    MTGA~\cite{mtga}                        & 63.90 & 86.38 & 75.08 \\
    MSTP++(+mixup)~\cite{mstpp}             & 64.30 & 85.86 & 75.06 \\
    HFR-Lip~\cite{Semantics-Aware}          & 66.47 & 87.67 & 77.06 \\
    STCNet~\cite{stcnet}                    & 65.86 & 87.43 & 76.62 \\
    \midrule
    \textbf{Ours(TVTA)}            & \shortstack[c]{67.23 (student)\\66.51 (teacher)} & \shortstack[c]{87.79 (student)\\87.19 (teacher)} & \shortstack[c]{77.49 (student)\\76.82 (teacher)}\\
    \bottomrule
  \end{tabular}
\end{table}

Table~\ref{tab:comparison} shows that the proposed method achieves an overall accuracy of 77.49\% with the student model, surpassing HFR-Lip by 0.43 points and STCNet by 0.87 points. On the challenging Acc1 subset, our method reaches 67.23\%, which is higher than most competing methods and remains close to the best previously reported result. On Acc2, the student achieves 87.79\%, which is also a particularly strong result among the compared methods. These results indicate that the proposed temporal aggregation strategy performs prominently across both confusing-word and common-word subsets. In the current experiment, the student achieves higher overall accuracy than the teacher, while both branches remain competitive.

\begin{figure*}[t]
  \centering
  \begin{minipage}[t]{0.49\textwidth}
    \centering
    \includegraphics[width=\linewidth]{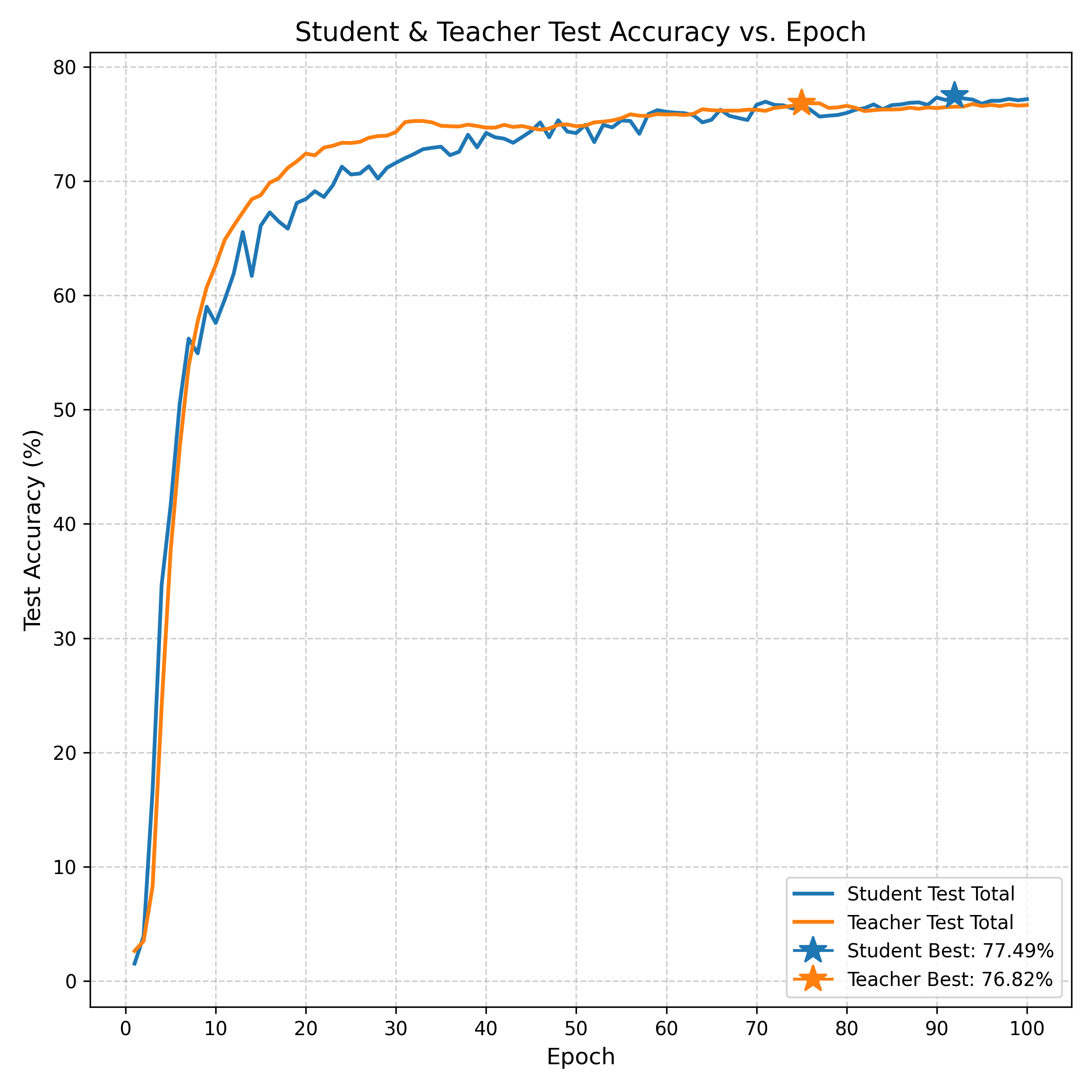}
  \end{minipage}\hfill
  \begin{minipage}[t]{0.49\textwidth}
    \centering
    \includegraphics[width=\linewidth]{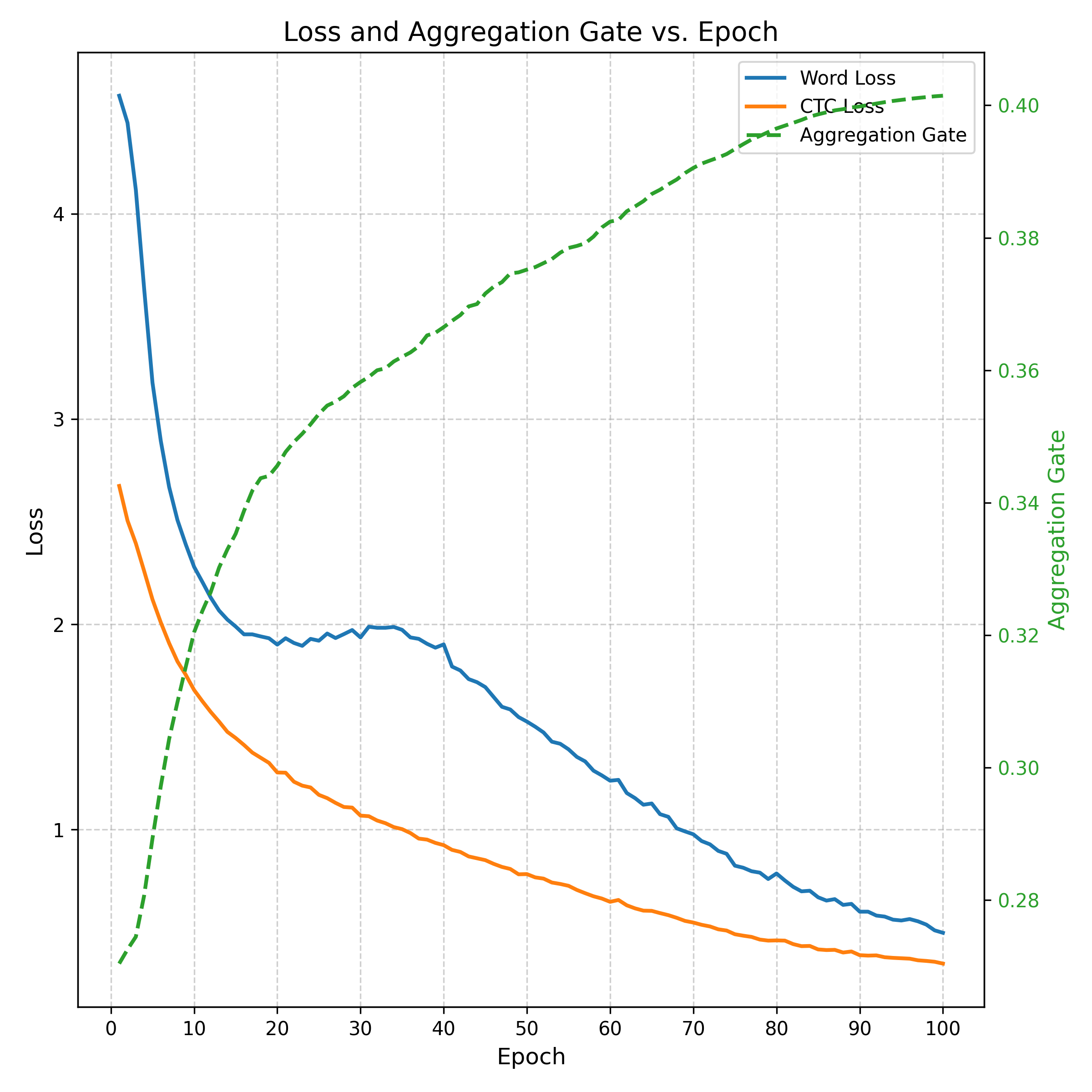}
  \end{minipage}
  \caption{Training dynamics of the proposed framework. Left: accuracy curves of the student and EMA teacher. Right: evolution of the word loss, CTC loss, and pool gate, which controls the interpolation between temporal mean pooling and the viseme-guided gated aggregation branch.}
  \Description{The left plot shows student and EMA teacher accuracy during training, with the teacher curve being smoother and stronger in the middle stage before converging with the student later. The right plot shows steadily decreasing word and CTC losses, together with an increasing pool gate that indicates growing reliance on the viseme-guided gated aggregation branch over mean pooling.}
  \label{fig:training_dynamics}
\end{figure*}

Fig.~\ref{fig:training_dynamics} provides additional evidence on the optimization behavior of the proposed framework. In the left panel, the EMA teacher curve is noticeably smoother than the student curve throughout training, indicating that parameter averaging effectively suppresses optimization fluctuations caused by strong perturbations. Moreover, the teacher performs better in the middle stage, while the two branches become comparable later and the student eventually attains the best final accuracy. This behavior supports the role of the EMA teacher as a stable regularizing target rather than the final inference branch.

The right panel further illustrates the effect of the proposed VGA. Both the word loss and the CTC loss decrease steadily, showing that the two supervision signals are optimized jointly without obvious conflict. More importantly, the aggregation gate increases progressively during training, indicating that the model gradually assigns more weight to the viseme-guided aggregation branch relative to the original mean-pooling path. This trend is consistent with the idea that the segment context produced by the CTC decoder becomes increasingly reliable as training proceeds. Overall, these dynamics suggest that the viseme-aware segment representation learned by the CTC decoder provides useful temporal cues for final word recognition, rather than acting only as an auxiliary loss.

\begin{figure}[t]
  \centering
  \includegraphics[width=\linewidth]{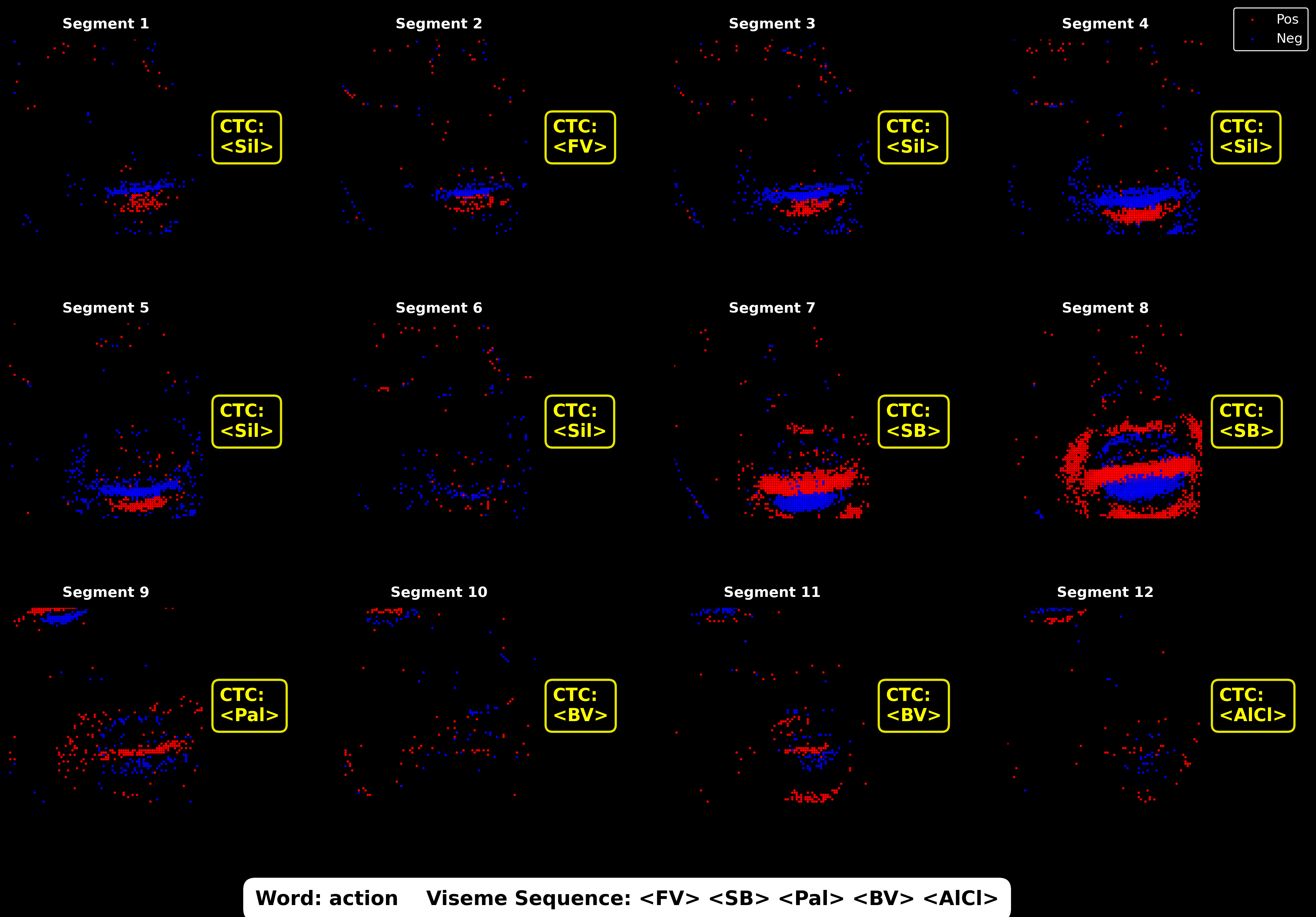}
  \caption{Qualitative visualization of VGA decoding on an event voxel sample of the word ``action''. The 12 visual segments are decoded into an ordered viseme sequence, where blank segments are skipped and the remaining outputs follow a consistent visual articulation progression.}
  \Description{A twelve-panel visualization of an event voxel sample for the word action. Each panel shows one visual segment with positive and negative events and the corresponding CTC prediction. Several segments are decoded as blank, while the effective decoded sequence follows a viseme progression consistent with the target word.}
  \label{fig:ctc_decode_vis}
\end{figure}

Fig.~\ref{fig:ctc_decode_vis} provides a qualitative example of the decoding behavior of VGA. For the word ``action'', the model partitions the event voxel into 12 visual segments and produces the ordered viseme sequence $\langle$FV$\rangle$ $\langle$SB$\rangle$ $\langle$Pal$\rangle$ $\langle$BV$\rangle$ $\langle$AICI$\rangle$ after removing blank predictions. Although the viseme representation is intentionally coarse, the decoded outputs still follow a consistent visual articulation progression of the target word. This result provides additional evidence that the CTC decoder learns meaningful viseme-aware structure rather than merely serving as an auxiliary optimization signal.

\subsection{Ablation Study}

\begin{table}[t]
  \centering
  \caption{Ablation study of the proposed modules.}
  \label{tab:ablation_modules}
  \renewcommand{\arraystretch}{1.25}
  \begin{tabular}{cccc c@{}}
    \toprule
     \textbf{DropPath} & \textbf{Teacher--Student} & \textbf{TDA} & \textbf{VGA}& \textbf{Acc (\%)} \\
    \midrule
                &            &            &            &  74.09\\
    \midrule
     \cmark &            &            &            &  74.96\\
    \midrule
     \cmark & \cmark &            &            &  \shortstack[c]{75.78 (student)\\75.82 (teacher)}\\
    \midrule
     \cmark & & \cmark &            &  75.59\\
    \midrule
     \cmark & & & \cmark &  76.58\\
    \midrule
 \cmark & \cmark & \cmark & &\shortstack[c]{76.54 (student)\\76.58 (teacher)}\\
    \midrule
 \cmark & \cmark & \cmark & \cmark &\shortstack[c]{77.49 (student)\\76.82 (teacher)}\\
 \midrule
  \end{tabular}
\end{table}

\begin{table*}[t]
  \centering
  \begin{minipage}[t]{0.31\textwidth}
    \centering
    \captionof{table}{CTC-to-CE weight ratio ablation.}
    \label{tab:ablation_ctc_ratio}
    \renewcommand{\arraystretch}{1.2}
    \begin{tabular}{@{}lc@{}}
      \toprule
      $\lambda_{\mathrm{ctc}} / \lambda_{\mathrm{ce}}$ & \textbf{Acc (\%)} \\
      \midrule
      0.5&  \shortstack[c]{76.34 (student)\\76.10 (teacher)}\\
      \midrule
      1.0&  \shortstack[c]{77.49 (student)\\76.82 (teacher)}\\
      \midrule
      2.0&  \shortstack[c]{76.22 (student)\\76.54 (teacher)}\\
      \bottomrule
    \end{tabular}
  \end{minipage}\hfill
  \begin{minipage}[t]{0.31\textwidth}
    \centering
    \captionof{table}{KL-to-CE weight ratio ablation.}
    \label{tab:ablation_kl_ratio}
    \renewcommand{\arraystretch}{1.2}
    \begin{tabular}{@{}lc@{}}
      \toprule
      $\lambda_{\mathrm{kl}} / \lambda_{\mathrm{ce}}$ & \textbf{Acc (\%)} \\
      \midrule
      0.5&  \shortstack[c]{76.38 (student)\\76.58 (teacher)}\\
      \midrule
      1.0&  \shortstack[c]{77.49 (student)\\76.82 (teacher)}\\
      \midrule
      2.0&   \shortstack[c]{76.84 (student)\\76.04 (teacher)}\\
      \midrule
    \end{tabular}
  \end{minipage}\hfill
  \begin{minipage}[t]{0.31\textwidth}
    \centering
    \captionof{table}{TDA temporal module ablation.}
    \label{tab:ablation_temporal_module}
    \renewcommand{\arraystretch}{1.2}
    \begin{tabular}{@{}lc@{}}
      \toprule
      \textbf{Temporal Module} & \textbf{Acc (\%)} \\
      \midrule
      Mamba &  \shortstack[c]{77.49 (student)\\76.82 (teacher)}\\
      \midrule
      GRU &  \shortstack[c]{75.82 (student)\\76.12 (teacher)}\\
      \midrule
      LSTM &  \shortstack[c]{76.68 (student)\\75.40 (teacher)}\\
      \bottomrule
    \end{tabular}
  \end{minipage}
\end{table*}

\begin{table*}[t]
  \centering
  \begin{minipage}[t]{0.48\textwidth}
    \centering
    \captionof{table}{VGA segment model ablation.}
    \label{tab:ablation_vga_segment_model}
    \renewcommand{\arraystretch}{1.2}
    \begin{tabular}{@{}lc@{}}
      \toprule
      \textbf{Segment Model} & \textbf{Acc (\%)} \\
      \midrule
      Mamba &  \shortstack[c]{77.37 (student)\\77.03 (teacher)}\\
      \midrule
      GRU &  \shortstack[c]{77.27 (student)\\77.07 (teacher)}\\
      \midrule
      LSTM &  \shortstack[c]{77.49 (student)\\76.82 (teacher)}\\
      \bottomrule
    \end{tabular}
  \end{minipage}\hfill
  \begin{minipage}[t]{0.48\textwidth}
    \centering
    \captionof{table}{VGA segment ratio ablation.}
    \label{tab:ablation_segment_ratio}
    \renewcommand{\arraystretch}{1.2}
    \begin{tabular}{@{}lc@{}}
      \toprule
      \textbf{Segment Ratio} & \textbf{Acc (\%)} \\
      \midrule
      $1/5$ &  \shortstack[c]{77.49 (student)\\76.82 (teacher)} \\
      \midrule
      $1/3$ &  \shortstack[c]{76.86 (student)\\76.48 (teacher)}\\
      \midrule
      $1/2$ &  \shortstack[c]{76.95 (student)\\76.24 (teacher)}\\
      \midrule
      $1/1$ &  \shortstack[c]{76.76 (student)\\75.98 (teacher)}\\
      \bottomrule
    \end{tabular}
  \end{minipage}
\end{table*}

Table~\ref{tab:ablation_modules} verifies the contribution of each component. Relative to the 74.09\% baseline, introducing DropPath improves accuracy to 74.96\%, indicating that moderate stochastic regularization is beneficial for event-based lip reading. Adding teacher--student training further raises the performance to 75.78\% for the student and 75.82\% for the teacher, confirming the effectiveness of perturbation-consistent supervision. Using TDA alone brings the student accuracy to 75.59\%, while introducing VGA alone yields a stronger result of 76.58\%, already surpassing the gains obtained by adding only TDA or only teacher--student training. When TDA is added on top of DropPath and teacher--student training, the student accuracy reaches 76.54\%, validating the benefit of temporal modeling before spatial compression. After further introducing VGA, the student attains the best result of 77.49\%, corresponding to a total gain of 3.40 points over the baseline. The standalone VGA result is already close to the combination of TDA and teacher--student training, highlighting the strong contribution of viseme-guided temporal aggregation.

Table~\ref{tab:ablation_ctc_ratio} examines the balance between the CTC loss and the cross-entropy loss. The best student accuracy of 77.49\% is achieved at $\lambda_{\mathrm{ctc}} / \lambda_{\mathrm{ce}}=1.0$. Reducing the ratio to 0.5 lowers the student accuracy to 76.34\%, while increasing it to 2.0 further decreases it to 76.22\%. This trend indicates that the CTC objective is most effective when kept at a balanced scale relative to word-level supervision; a weak CTC signal provides insufficient temporal guidance, whereas an overly strong one biases optimization toward the auxiliary viseme task.

Table~\ref{tab:ablation_kl_ratio} evaluates the weight of KL consistency regularization. The best student accuracy of 77.49\% is achieved at $\lambda_{\mathrm{kl}} / \lambda_{\mathrm{ce}}=1.0$. Reducing the ratio to 0.5 decreases the student accuracy to 76.38\%, while increasing it to 2.0 also leads to a lower result of 76.84\%. These results suggest that a balanced consistency weight is most effective in the current setting, whereas either weaker or stronger consistency regularization leads to inferior student performance. For the teacher branch, the result is also strongest at $\lambda_{\mathrm{kl}} / \lambda_{\mathrm{ce}}=1.0$.

Table~\ref{tab:ablation_temporal_module} compares different temporal modeling modules used inside TDA. Replacing Mamba with GRU leads to a clear drop in student accuracy from 77.49\% to 75.82\%, while LSTM achieves 76.68\%, which is better than GRU but still below Mamba. The same advantage is also reflected in the teacher results, where Mamba obtains 76.82\%, outperforming GRU by 0.70 points and LSTM by 1.42 points. These results confirm that the selective state-space modeling mechanism in Mamba is more effective than recurrent alternatives for capturing local temporal dynamics before spatial aggregation, which directly supports the design motivation of TDA.

Table~\ref{tab:ablation_vga_segment_model} compares different segment modeling modules used inside VGA. All three alternatives produce competitive and very close results: LSTM yields the best student accuracy of 77.49\%, only slightly higher than Mamba at 77.37\% and GRU at 77.27\%, while for the teacher branch GRU reaches the highest value of 77.07\% and Mamba and LSTM remain close. These results more clearly indicate that the Segment Model in VGA is relatively insensitive to the specific temporal module choice, as different sequence modeling units lead to only marginal performance differences. Compared with the temporal modeling stage in TDA, the segment modeling branch in VGA exhibits substantially lower sensitivity to module type. Under the current experimental setting, LSTM provides the most favorable overall trade-off.

Table~\ref{tab:ablation_segment_ratio} studies the temporal segment ratio used in VGA. The best result is obtained at the $1/5$ setting, which achieves 77.49\% for the student and 76.82\% for the teacher. When the segment ratio is increased to $1/3$ or $1/2$, the performance drops modestly, and it decreases further at $1/1$. A plausible explanation is that overly fine segmentation introduces too many sparse or nearly blank segments in event streams, which weakens the stability of segment-level modeling and reduces the usefulness of the decoded segment context. This observation indicates that a moderate segment granularity is more suitable for viseme-guided aggregation than excessively dense partitioning.

\section{Conclusion}

In this paper, we presented a temporally enhanced framework for event-based lip reading that addresses the limitation of performing temporal modeling only after spatial compression. To better preserve localized dynamic cues in sparse event streams, we introduced Trajectory-Aware Differential Aggregation, which performs temporal modeling at each spatial location before adaptive spatial aggregation. On top of this representation, we further proposed Viseme-Guided Aggregation, which combines a CTC decoder and a viseme-guided gated aggregation branch to inject viseme-aware sequence supervision and improve final temporal aggregation. In addition, an EMA teacher--student training strategy was adopted to enhance robustness under strong event perturbations.

Extensive experiments on the DVS-Lip dataset demonstrated the effectiveness of the proposed design. Our method achieved an overall accuracy of 77.49\%, surpassing previous state-of-the-art methods. The ablation studies further verified that TDA, VGA, and teacher--student consistency all contribute positively to performance, while the temporal-module comparison confirmed the advantage of Mamba over GRU and LSTM for local temporal modeling inside TDA.

In future work, we plan to extend the proposed framework to more challenging continuous or sentence-level event-based visual speech recognition settings.

\bibliographystyle{ACM-Reference-Format}
\bibliography{references}

\end{document}